\documentclass[runningheads]{llncs}
\usepackage{hyperref}
\hypersetup{
    colorlinks=true,
    linkcolor=red,
    filecolor=green,      
    urlcolor=blue,
    citecolor=green,
}

\usepackage[T1]{fontenc}
\usepackage{graphicx}
\usepackage{amsmath}
\usepackage{cite}
\usepackage{textcomp}
\usepackage{xcolor}
\usepackage{booktabs} 
\usepackage{color}
\usepackage{colortbl}  
\usepackage{hyperref}
\usepackage{subcaption}
\usepackage{multirow}
\usepackage{bbding} 
\newcommand{\smallerfonttable}{\fontsize{6pt}{7pt}\selectfont}
\usepackage{float} 
\begin{document}
\title{LLDif: Diffusion Models for Low-light Emotion Recognition}
%
%
\author{Zhifeng Wang\inst{1} \and
Kaihao Zhang\inst{2} \and
Ramesh Sankaranarayana\inst{1}}
\authorrunning{Z.Wang et al.}
%
\institute{College of Engineering and Computer Science, Australian National University, Canberra, ACT, Australia  \and
School of Computer Science and Technology, Harbin Institute of Technology, Shenzhen, China\\
}
\maketitle              
\begin{abstract}
This paper introduces LLDif, a novel diffusion-based facial expression recognition (FER) framework tailored for extremely low-light (LL) environments. Images captured under such conditions often suffer from low brightness and significantly reduced contrast, presenting challenges to conventional methods. These challenges include poor image quality that can significantly reduce the accuracy of emotion recognition. LLDif addresses these issues with a novel two-stage training process that combines a Label-aware CLIP (LA-CLIP), an embedding prior network (PNET), and a transformer-based network adept at handling the noise of low-light images. The first stage involves LA-CLIP generating a joint embedding prior distribution (EPD) to guide the LLformer in label recovery. In the second stage, the diffusion model (DM) refines the EPD inference, ultilising the compactness of EPD for precise predictions. Experimental evaluations on various LL-FER datasets have shown that LLDif achieves competitive performance, underscoring its potential to enhance FER applications in challenging lighting conditions.
\end{abstract}
\section{Introduction}
\begin{figure}[t]
\centering
\includegraphics[width=0.95\linewidth]{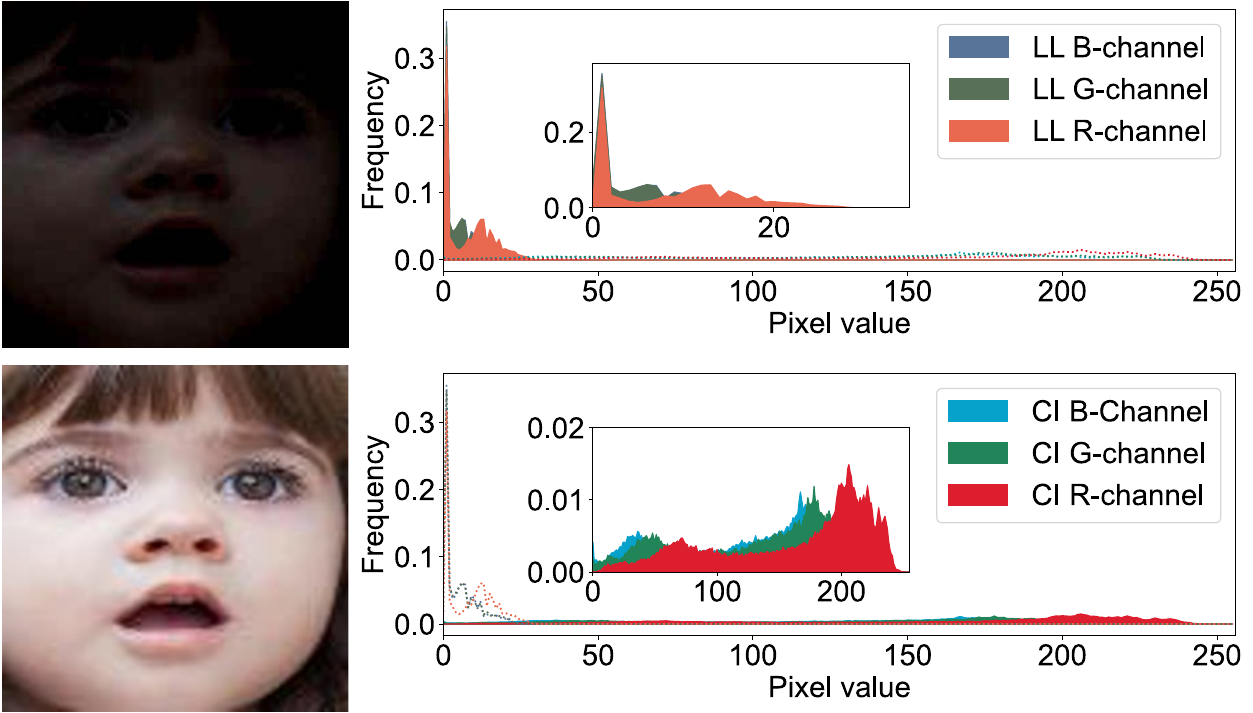}
\caption{Top: the low-light image (LL) shows a child's face that is shadowed and details are obscured, making it challenging to discern fine facial expressions. Bottom: In the normal-light image (CI) at the bottom, the child's face is clearly visible with good detail, essential for recognizing emotions. }
\label{low_light_clear_images_histogram}
\end{figure}

In the domain of computer vision, precisely identifying facial emotions presents a notable challenge, particularly in extremely low-light environments. Such environments can significantly impair the quality of captured images, leading to degraded visibility of facial features, which are crucial for precise emotion recognition. This degradation not only destroys the basic structure of the face but also introduces noise and distortion, further complicating the task for emotion recognition algorithms. In Fig. \ref{low_light_clear_images_histogram}, the low-light image (LL) at the top shows a child's face that is shadowed and details are obscured, making it challenging to discern fine facial expressions. The histograms indicate that most pixel values are clustered toward the darker end of the spectrum, which suggests limited brightness and contrast in the image. In the normal-light image (CI) at the bottom, the child's face is clearly visible with good detail, essential for recognizing emotions. The histograms show a more even distribution of pixel values across the spectrum, with higher frequencies in the mid to high ranges, indicating better brightness and contrast. Traditional facial expression recognition (FER) methods \cite{2021arm,2021dmue,2021manet,zhang2017facial,niu2022four,wang2024htnet} perform well under normal-light conditions; however, their effectiveness is considerably diminished in low-light scenarios due to the loss of subtle facial structures. There is a need for robust methodologies that can overcome the challenges posed by low brightness while maintaining high accuracy in emotion recognition.

Currently, several approaches have been developed to tackle the challenge of learning from noisy data in the field of emotion recognition. RUL \cite{2021rul} proposes to improve facial expression recognition by weighing uncertainties based on the difficulty of samples to enhance performance in noisy environments. SCN \cite{2020scn} addresses uncertainties in facial expression recognition efforts by using a self-attention block to choose training samples and correcting uncertain labels by using a relabeling approach, thereby improving the learning process's dependability. However, both methods require relabeling the samples based on the samples' difficulties. EAC \cite{2022eac} addresses noisy labels by using flipped image consistency and selective features, preventing the model from relying on misleading features and thereby improving learning accuracy. However, when these techniques are used in low-light images, they encounter challenges. In particular, RUL \cite{2021rul} and EAC \cite{2022eac} are based on the assumption of minimal losses. In extremely low-light settings, where clear, fine facial details are lacking, these approaches might mistakenly equate challenging samples with noisy ones since both can display high loss values in the training of low-light images.
\begin{figure*}[t!]
\centering
\includegraphics[width=0.98\linewidth]{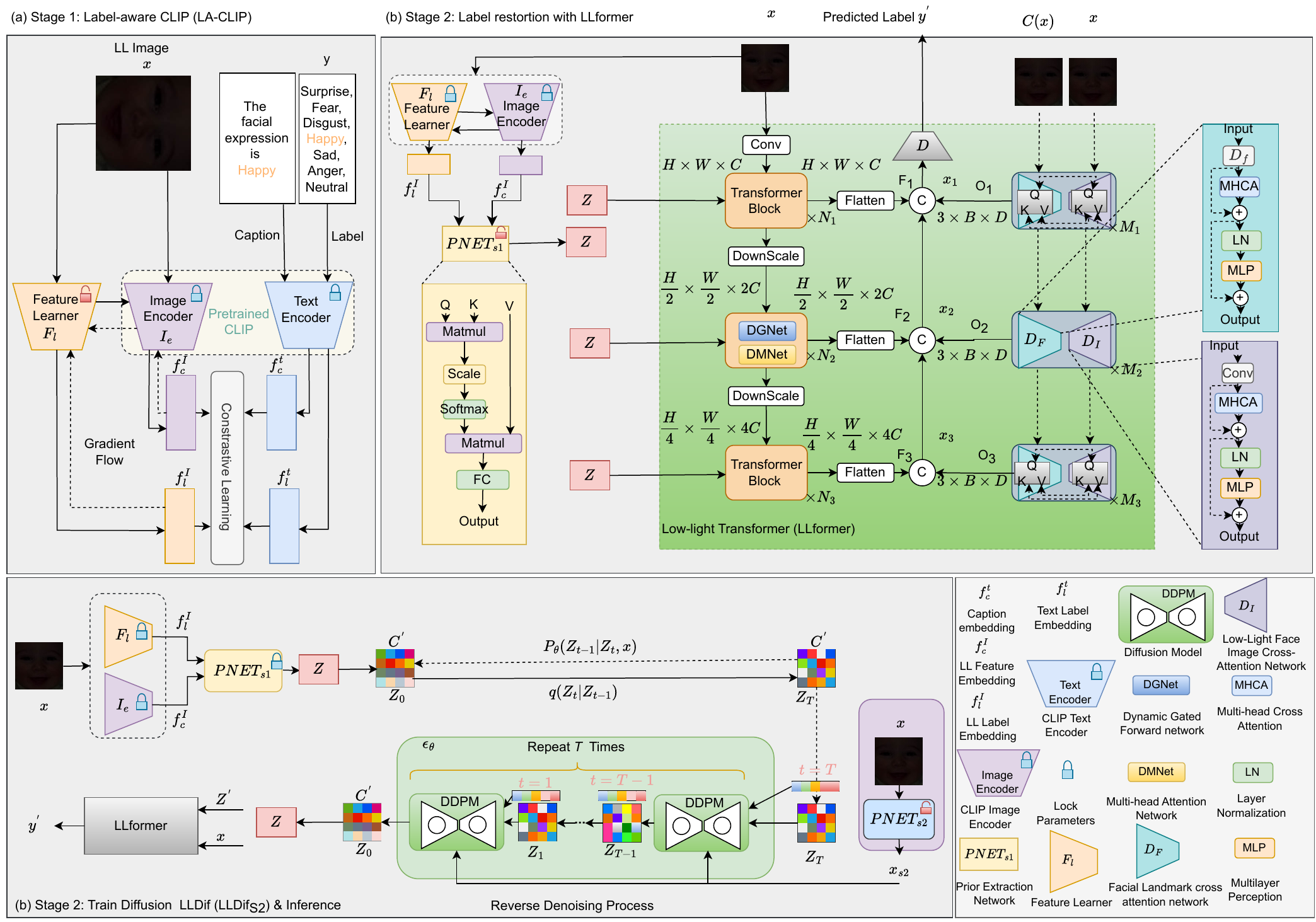}
\caption{The proposed LLDif framework, comprising Label-aware CLIP (LA-CLIP), LLformer, PNET, and a denoising network. LLDif employs a two-stage training method: (1) Initially, we apply LA-CLIP to process the low-light image alongside its image caption and label, producing a Joint Embedding Prior Distribution (EPD) Z. This EPD is then used to instruct the LLformer in label restoration. (2) During the second stage, the diffusion model (DM) undergoes training to directly deduce the precise Embedding Prior Distribution (EPD) from images captured in low-light conditions.}
\label{LRDif_architecture}
\end{figure*}

To solve these issues, this paper proposes a novel method for handling noisy images in low-light conditions, departing from the conventional method of identifying noisy samples by their loss values. Instead, we introduce a distinctive approach centered on learning the joint distribution of noise labels and images via feature extraction and label restoration. We aim to create a diffusion-based network for facial expression recognition (FER) that use the capabilities of diffusion models (DMs) for effective label restoration by aligning them with their related images. To achieve this, we present LLDif. Considering the transformer's capability to handle long-range pixel dependencies, we employ transformers as the foundational blocks of the LLDif architecture. We organize transformer blocks in a U-Net configuration to form the Low-Light Transformer (LLformer), which is aimed at extracting features at multiple levels. The LLformer comprises two parallel networks: the DTNet, tasked with extracting latent features from low-light images at various depths, and the DLNet, which focuses on identifying the similarities between low-light images and facial landmarks. LLDif adopts a two-stage training approach: (1) In the first stage, as illustrated in Fig. \ref{LRDif_architecture} (a), we use LA-CLIP to process the low-light image along with its image caption and label, generating a Joint Embedding Prior Distribution (EPD) Z. This EPD is then utilized to guide the LLformer in label restoration. (2) In the second stage, shown in Fig. \ref{LRDif_architecture} (b), the diffusion model (DM) can be trained to deduce the accurate EPD directly from low-light images. Owing to the compactness of EPD Z, the DM can make highly accurate EPD predictions, achieving consistent high accuracy after only a few iterations.

This study offers several notable contributions, detailed as follows: 1) We introduce a innovative diffusion-based approach designed to address the challenges encountered in facial expression recognition, particularly those arising from diminished brightness and contrast in low-light conditions. 2) Our LLDif model harnesses the powerful distribution mapping capabilities of diffusion models (DMs) to generate an accurate embedding prior distribution (EPD), significantly enhancing the precision and reliability of facial expression recognition (FER) results. This method stands out for its independence from the need to understand the dataset's uncertainty distribution, distinguishing it from prior approaches. 3) Extensive testing has demonstrated that LLDif achieves impressive performance in emotion recognition tasks across three low-light FER datasets, underscoring its effectiveness.
\section{Related Work}

\textbf{Facial Expression Recognition}. Facial Expression Recognition (FER)\cite{zhang2023geometric,wang2024lrdif, zhang2024authentic} focuses on enabling computers to interact with humans by identifying human facial expressions. In recent years, the accuracy of recognizing expressions under normal-light conditions has seen substantial improvements. Kollias \textit{et al.} \cite{kollias2020exploiting} introduces a CNN-RNN hybrid method that leverages multi-level visual features for dimensional emotion recognition. Zhao \textit{et al.} \cite{zhao2021former} introduces Former-DFER, a dynamic transformer that combines spatial and temporal transformers to robustly capture facial features against occlusions and pose variations, achieving top performance on an emotion recognition dataset. The Expression Snippet Transformer (EST) \cite{liu2023expression} enhances video-based facial expression recognition by decomposing videos into expression snippets for detailed intra- and inter-snippet analysis, significantly outperforming conventional CNN-based approaches. Vazquez \textit{et al.} \cite{vazquez2022transformer} introduces a Transformer-based model, pre-trained on unlabeled ECG datasets and fine-tuned on the AMIGOS dataset, achieving top emotion recognition performance by leveraging attention mechanisms to emphasize relevant signal parts.

\textbf{Diffusion Models}. Diffusion models are now utilized across a wide range of tasks, including image enhancement for higher resolution, as mentioned by Shang \textit{et al.} (2024) \cite{shang2024resdiff}, and creative image modifications, as highlighted by Yang \textit{et al.} (2023) \cite{yang2023paint}. Moreover, the latent features captured by diffusion models have proven beneficial for classification tasks such as image classification, as noted by Han \textit{et al.} (2022) \cite{han2022card}, and for segmentation in medical imaging, as demonstrated by Wu \textit{et al.} (2024) \cite{wu2024medsegdiff}. Zhang \textit{et al.} \cite{zhang2023sine} introduces a novel approach for editing single images using pre-trained diffusion models, combining model-based guidance with patch-based fine-tuning to prevent overfitting and enable high-resolution content creation and manipulation based on textual descriptions. Rahman \textit{et al.} \cite{rahman2023ambiguous} presents a diffusion model-based approach for medical image segmentation that learns from collective expert insights to generate a variety of accurate segmentation masks, outperforming existing models in capturing natural variations and evaluated by a new metric aligned with clinical standards.
\section{Methods}
\subsection{Label-aware CLIP}
The key idea of LA-CLIP is to train the feature learner $F_l$ to output low-light features while simultaneously predicting the image's label. As summarized in Fig. \ref{LRDif_architecture} (a), the low-light feature embedding $f_c^{I}$ is matched with the image's caption $f_c^{t}$. Moreover, the low-light label embedding $f_l^{I}$, predicted by the feature learner $F_l$, is aligned with the input label embedding $f_l^{t}$. This module helps to create embeddings that correlate visual features with textual annotations, which could be vital for low-light emotion recognition. It is designed to support the LLformer in label restoration, leveraging pre-trained models to guide the network in accurately predicting labels for low-light images.

As depicted in the yellow box of Fig. \ref{LRDif_architecture} during stage 2, $PNET_{s1}$ employs cross-attention layers to infer the Embedding Prior Distribution (EPD) Z. Following this extraction, DTNet leverages the EPD to aid in label recovery. Within DTNet, as shown in the same yellow box of Fig. \ref{LRDif_architecture}, the architecture comprises DMNet and DGNet. We use the pre-trained LA-CLIP model to get the low-light feature embedding $f_c^{l}$ and low-light label embedding $f_l^{I}$; these embeddings are then input into PNET${s1}$. The output from PNET${s1}$ is the EPD Z, denoted as $Z \in R^C$. This process is detailed in (Eq. \ref{Z_FPEN_S1}):
\begin{equation}
   Z = PNET_{S1}(F_l(x),I_e(x)).
   \label{Z_FPEN_S1}
\end{equation}

\begin{figure}[h]
\centering
\includegraphics[width=0.5\linewidth]{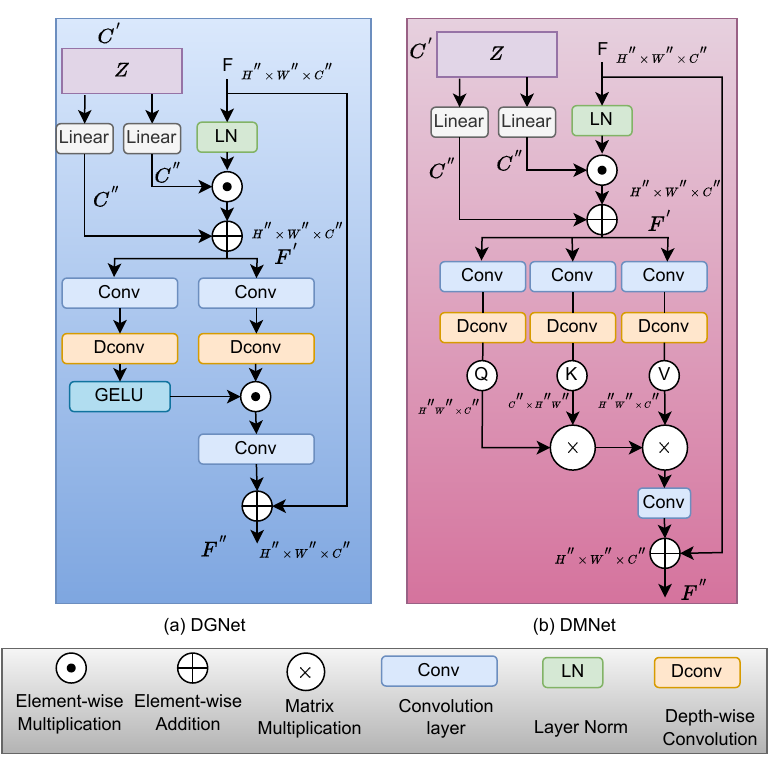}
\caption{The overview of DTNet, which consists of DGNet and DMNet.}
\label{DTNet}
\end{figure}

Subsequently, $Z$ is fed into the DTNet in Fig.\ref{DTNet}, acting as adjustable parameters to support the process of label restoration, as detailed in Equation (\ref{F_w1_ln}).
\begin{equation}
F^{'} = W_{1}^{l}Z\circ LN(F) +  W_{2}^{l}Z,
\label{F_w1_ln}
\end{equation}
here, \(W\) represents the weights of a fully connected layer, $LN$ denotes layer normalization and \(\circ\) symbolizes element-wise multiplication.
In DMNet Fig.\ref{DTNet} (b), we process the entire image to extract detailted information. The features \(F'\) are converted into three different vectors: key \(K\), query \(Q\), and value \(V\), through a convolutional layer. These vectors are reshaped as \(Q\) to \(R^{H''W''\times C''}\), \(K\) to \(R^{C''\times H''W''}\), and \(V\) to \(R^{H''W''\times C''}\), making them compatible for subsequent operations. By multiplying \(Q\) and \(K\), the model can identify which image regions to focus on, and generate an attention map \(A \in R^{C''\times C''}\). This operation in DMNet is depicted in the following equation Equation (\ref{f_wc_v_softmax}):
\begin{equation}
F^{"} = W_{c}V\times softmax(K\times Q/\alpha) + F,
\label{f_wc_v_softmax}
\end{equation}
where \(\alpha\) serves as a tunable parameter during the training phase. Following this, the DGNet focuses on extracting both local and neighboring features through aggregation. This is achieved by employing a small Convolution (\(1\times1\)) to extract local features, and a larger Convolution (\(3\times3\)) to collect information from adjacent pixels. Furthermore, a specialized gating mechanism is utilized to ensure only the most important information is captured. The entire process within DGNet is depicted in the following equation (Eq. (\ref{f_gelu_w1_w1c})):
\begin{equation}
F^{"} = GELU(W^{1}_{d}W^{1}_{c}F^{'})\circ W^{2}_{d}W^{2}_{c}F^{'} + F.
\label{f_gelu_w1_w1c}
\end{equation}
\subsection{Dynamic Landmarks and Image Network (DLNet)}
Within the DLNet, a cross window attention approach is utilized to process features from both 2D facial landmarks and related images taken in low-light conditions. We start by dividing the low-light image features, denoted as \(X_{ll} \in R^{N\times D}\), into various distinct, non-overlapping windows \(x_{ll}\in R^{M\times D}\). In parallel, features from facial landmarks, represented as \(X_{fl} \in R^{C\times H\times W}\), are downscaled to align with the dimensions of these windows, yielding \(x_{fl} \in R^{c\times h\times w}\), where the dimension \(c\) matches \(D\) and the production \(h\) and \(w\) equate to \(M\). This setup enables the application of cross-attention between features of facial landmarks and low-light images, as depicted in Equation (Eq. \ref{Q_O_O_FLM}).
\begin{align}
Q   &=x_{fl}w_{Q}, K = x_{ll}w_{K}, V = x_{ll}w_{V},\\
O_{i}&= Softmax(\frac{Q_{i}K_{i}^T}{\sqrt{d}} +b)V_{i}, i=1,...,N,\\
O &= [O_{1},O_{2},...,O_{N}]W_{O},
\label{Q_O_O_FLM}
\end{align}
where \(w_{O}\), \(w_{K}\), \(w_{Q}\) and \(w_{V}\) represent the weight matrices, and \(b\) denotes the corresponding positional bias.

This cross-attention mechanism is implemented on every window of the low-light image, termed as MHCA. The equations that describe the transformer encoder within LLDif are presented as follows (Eq. (\ref{x-ll-mhca})):
\begin{align}
X^{'}_{ll} &= MHCA(X_{ll}) +X_{ll},\\
X^{"}_{ll} &= MLP(LN(X^{'}_{ll})) + X^{'}_{ll},
\label{x-ll-mhca}
\end{align}
the fusion of output features \(F\) from DTNet and \(O\) from DLNet is required to produce the combined multi-scale features \(x_1\), \(x_2\), and \(x_3\). This involves concatenating the corresponding features: \(x_1 = Concat(F_1,O_1)\), \(x_2 = Concat(F_2,O_2)\), and \(x_3 = Concat(F_3,O_3)\). Following this, the fused features \(X\) undergo additional processing through standard transformer blocks.
\begin{align}
X &=[x_1,x_2,x_3],\\
X^{'} &=MSA(X) +X,\\
y^{'} &=MLP(LN(X^{'})) +X^{'},
\end{align}
where \( MSA \) denotes the self-attention blocks with multiple heads and \( LN \) refers to the layer normalization. The definition of the training loss is given as follows (Eq. (\ref{l_ce_loss})):
\begin{equation}
\mathcal{L}_{ce} = - \sum_{i=1}^{N} \sum_{c=1}^{M} y_{ic} \log(p_{ic}).
\label{l_ce_loss}
\end{equation}
Our model is trained using the cross-entropy loss function, where \(M\) is the number of distinct classes, and \(N\) signifies the total count of samples. Here, \(y_{ic}\) indicates whether class \(c\) is the correct classification for observation $i$, and \(p_{ic}\) is the probability predicted by the model.

\subsection{Diffusion Model for Label Restoration}
In the second stage, as shown in Fig. \ref{LRDif_architecture} (b), the diffusion model's (DM) strong capabilities are employed to approximate the joint Embedding Prior Distribution (EPD). Initially, the pre-trained LA-CLIP and PNET$_{S1}$ is used to acquire the EPD \(Z \in R^{C}\). Following this, the diffusion technique is applied to \(Z\), resulting in a generated sample \(Z_T \in R^{C}\), as explained in (Eq. (\ref{sample_z})):
\begin{equation}
q(Z_T | Z) = \mathcal{N}(Z_T; \sqrt{\bar{\alpha}_T} Z, (1 - \bar{\alpha}_T)I).
\label{sample_z}
\end{equation}
here, \(T\) represents the total count of diffusion steps. The variable \(\alpha_{t}\) is defined as \(1-\beta_{t}\), and \(\bar{\alpha}_T\) denotes the cumulative product of \(\alpha_i\) for all steps from 0 to \(T\). The term \(\beta_{t}\) is a predetermined hyper-parameter, while \(\mathcal{N}(.)\) signifies the standard Gaussian distribution.

During the reverse process of the diffusion model, low-light images \(x\) are fed into PNET\(_{s2}\) to derive a conditional vector \(x_{s2} \in R^{C}\) as outlined in Equation (\ref{x_s2_fpen}).  
\begin{equation}
x_{s2} = \text{PNET}_{s2}(x),
\label{x_s2_fpen}
\end{equation}
where $\text{PNET}_{s2}$ includes convolutional layer, residual layer and linear layer, which will ensure the output's dimension of $\text{PNET}_{s2}$ is same as   $\text{PNET}_{s1}$.

The denoising network, represented as \(\epsilon_{\theta}\), estimate the noise for each specific time step \(t\). It processes the current noisy data \(Z_{t}^{'}\), the time step \(t\), and a conditional vector \(x_{s2}\), which is obtained from the low-light image via the stage-two prior distribution network PNET\(_{s2}\). The estimated noise, expressed as \(\epsilon_{\theta}(\text{Concat}(Z_{t}^{'},t,x_{s2}))\), is then utilized in the subsequent formula to determine the denoised data \(Z_{t-1}^{'}\) for the upcoming step, as illustrated in Equation (\ref{reverse_noise_process}):
\begin{equation}
Z_{t-1}^{'}  = \frac{1}{\sqrt{\alpha_t}}(Z_{t}^{'} - \epsilon_{\theta}(\text{Concat}(Z_{t}^{'},t,x_{S2}))\frac{1-\alpha_{t}}{\sqrt{1-\alpha_{t}}}).
\label{reverse_noise_process}
\end{equation}
After \(T\) iterations, we get the final embedding prior distribution (EPD), symbolized as \(Z_0'\). The stage-two prior distribution network (PNET\(_{s2}\)), together with the denoising network and the Low-Light Transformer (LLformer), are jointly optimized through the total loss function \(\mathcal{L}_{total}\), as depicted in Equation (\ref{l_total_loss}).
\begin{equation}
\mathcal{L}_{kl} = \sum_{i=1}^{C} Z_{\text{norm}}(i) \log(\frac{Z_{\text{norm}}(i)}{\bar{Z}_{\text{norm}}(i)}),
\label{l_kl_loss}
\end{equation}
\begin{equation}
\mathcal{L}_{total} = \mathcal{L}_{ce} + \mathcal{L}_{kl} .
\label{l_total_loss}
\end{equation}
In this formula, \(Z_{\text{norm}}(i)\) and \(\bar{Z}_{\text{norm}}(i)\) refer to the EPDs derived from LA-CLIP and LLDif\(_{S2}\), respectively, both normalized through softmax. The term \(\mathcal{L}_{kl}\) represents a form of the Kullback-Leibler divergence, computed over C dimensions. The total loss, \(\mathcal{L}_{total}\), is formulated by adding the Kullback-Leibler divergence loss \(\mathcal{L}_{kl}\) (Eq. \ref{l_kl_loss}) to the Cross-Entropy loss \(\mathcal{L}_{ce}\) (Eq. \ref{l_ce_loss}). Since the EPD includes features from the low-light image and the corresponding emotion label encoded via a pretrained LA-CLIP model, LLDif's second stage (LLDif\(_{s2}\)) can provide accurate estimation for low-light image's label in a few steps. Notably, during the inference stage, LLDif doesn't need actual ground truth labels in the reverse diffusion process of DM.

\begin{table*}[t]
\centering
\smallerfonttable 
\renewcommand{\arraystretch}{0.7} 
\caption{Evaluation of accuracy (\%) compared to SOTA FER methods on RAF-DB, KDEF and FERPlus.}
\label{three-datasets-sota-comparison}
\begin{tabular}{lc|lc|lcc}
\toprule 
RAF-DB &   & FERPlus &  & KDEF &  \\
\midrule 
Methods & Acc. (\%) & Methods & Acc. (\%)  & Methods & Acc. (\%) \\
\midrule 
ARM\cite{2021arm}     & 90.42  &DACL\cite{2021dacl}  & 83.52 & DACL\cite{2021dacl} &88.61  \\
POSTER++\cite{2023posterv2} & \textcolor{red}{92.21} & POSTER++\cite{2023posterv2} & \textcolor{blue}{86.46} & POSTER++\cite{2023posterv2} & \textcolor{blue}{94.44} \\
RUL\cite{2021rul}     & 88.98  & RUL\cite{2021rul}  & 85.00  & RUL\cite{2021rul} & 87.83\\
DAN\cite{2023dan}     & 89.70  & DAN\cite{2023dan}  & 85.48       & DAN\cite{2023dan} &  88.77\\
SCN\cite{2020scn}     & 87.03  & SCN\cite{2020scn}  & 83.11  & SCN\cite{2020scn} & 89.55 \\
EAC\cite{2022eac}     & 90.35  & EAC\cite{2022eac}  & 86.18  & EAC\cite{2022eac} & 72.32\\
MANet\cite{2021manet} & 88.42  & MANet\cite{2021manet}&85.49 &MANet\cite{2021manet} & 91.75 \\
\midrule 
$Ours$     & \textcolor{blue}{91.72} &$Ours$   &\textcolor{red}{87.19}  &$Ours$   & \textcolor{red}{95.83}\\
\bottomrule 
\end{tabular}
\end{table*}
\begin{table}[t]
\centering
\smallerfonttable 
\setlength{\tabcolsep}{1pt} 
\caption{Evaluation of accuracy (\%) compared to SOTA FER methods on the LL-RAF-DB Dataset.}
\label{udc-raf-db-sota-comparison}
\begin{tabular}{lcccccccc}
\toprule 
\parbox{0pt}{\rule{0pt}{5ex}} &  DAN\cite{2023dan} & POSTER++\cite{2023posterv2} & EAC\cite{2022eac} & MANet\cite{2021manet} & RUL\cite{2021rul}  & SCN\cite{2020scn} &DACL \cite{2021dacl} &$Ours$ \\
\midrule 
 Acc.(\%)  & 79.27 & \textcolor{blue}{80.76} &78.72 & 78.45 & 77.57 & 75.20 & 75.68& \textcolor{red}{82.26}\\
\bottomrule 
\end{tabular}
\end{table}
\begin{figure}[tp]
\centering
\includegraphics[width=0.95\linewidth]{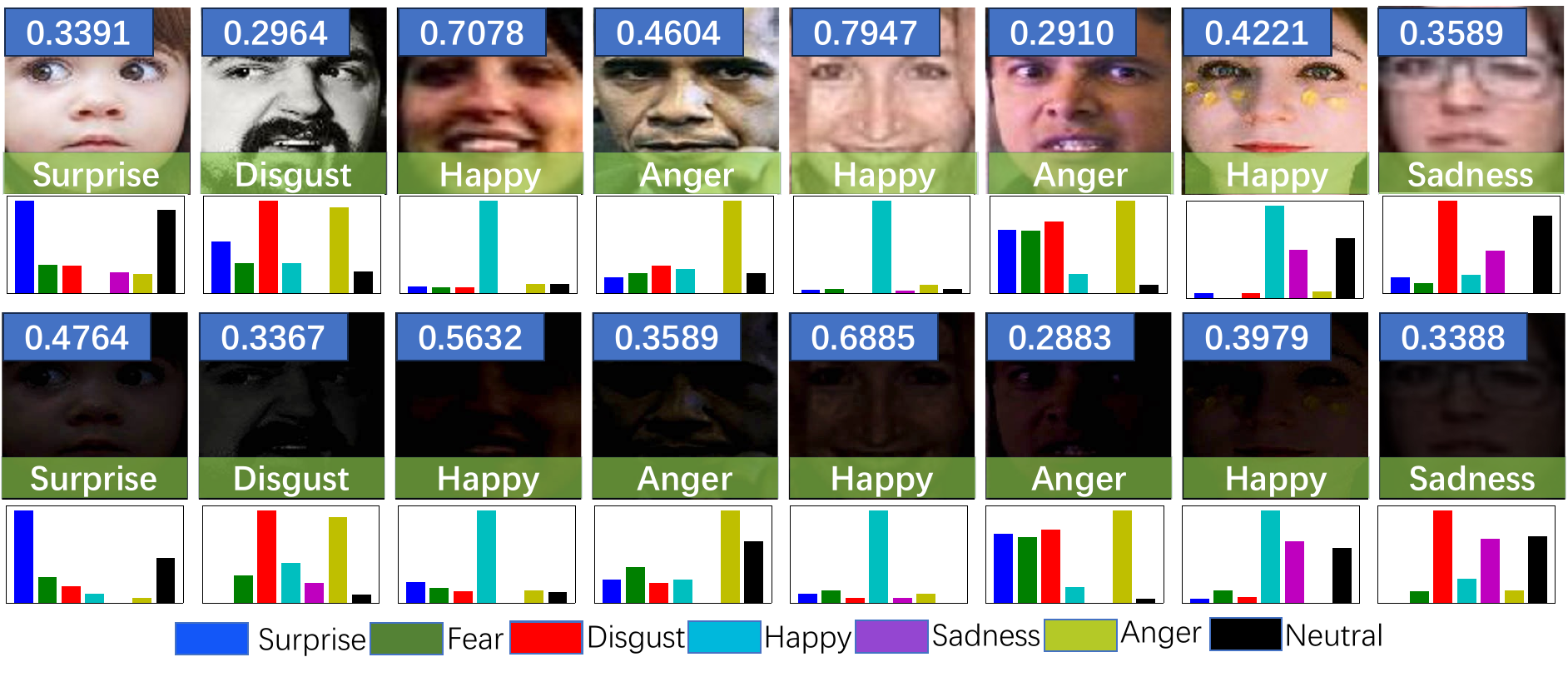}
\caption{Emotion distribution for samples in LL-RAF-DB dataset and RAF-DB.}
\label{probability-distribution-ours}
\end{figure}
\begin{table}[t]
\centering
\smallerfonttable 
\setlength{\tabcolsep}{1pt} 
\renewcommand{\arraystretch}{1} 
\caption{Evaluation of accuracy (\%) compared to SOTA FER methods on the LL-FERPlus Dataset.}
\label{udc-ferplus-sota-comparison}
\begin{tabular}{lcccccccc}
\toprule 
\parbox{0pt}{\rule{0pt}{5ex}} &  DAN\cite{2023dan} & POSTER++\cite{2023posterv2} & EAC\cite{2022eac} & MANet\cite{2021manet} & RUL\cite{2021rul}  & SCN\cite{2020scn} &DACL \cite{2021dacl} &$Ours$ \\
\midrule 
 Acc.(\%)  & 80.97 & \textcolor{blue}{81.44} & 80.46 & 80.34  & 79.35 & 74.95&77.05 & \textcolor{red}{82.25}\\
 
\bottomrule 
\end{tabular}
\end{table}
\begin{table}[ht]
\centering
\smallerfonttable 
\setlength{\tabcolsep}{1pt} 
\renewcommand{\arraystretch}{1} 
\caption{Evaluation of accuracy (\%) compared to SOTA FER methods on the LL-KDEF Dataset.}
\label{udc-kdef-sota-comparison}
\begin{tabular}{lcccccccc}
\toprule 
\parbox{0pt}{\rule{0pt}{5ex}} &  DAN\cite{2023dan} & POSTER++\cite{2023posterv2} & EAC\cite{2022eac} & MANet\cite{2021manet} & RUL\cite{2021rul}  & SCN\cite{2020scn} &DACL \cite{2021dacl} &$Ours$ \\
\midrule 
 Acc.(\%)  & 82.03 &  \textcolor{blue}{88.93} &  43.53  &  83.13  & 83.99 & 77.50 &86.69 & \textcolor{red}{92.97}\\
\bottomrule 
\end{tabular}
\end{table}
\section{Experiments}
\subsection{Datasets}
\textbf{LL-RAF-DB} dataset includes 12,271 images in the training set and 3,068 images in the testing set, offering a robust basis for assessing facial expression recognition (FER) algorithms under low-light conditions. Likewise, the RAF-DB dataset \cite{rafdb} includes 7 emotional categories and mirrors the testing and training configuration of LL-RAF-DB dataset. The expression distribution is consistent across both datasets.

\textbf{LL-FERPlus} dataset expands the scope to low-light conditions, presenting a comprehensive collection of 7,178 for testing and 28,709 images for training in low-light settings. The FERPlus dataset \cite{ferplus}, an extension of the FER2013 dataset, is enriched with additional labels from ten different annotators and features the same quantity of training and testing images as the LL-FERPlus dataset.

\textbf{LL-KDEF} dataset contains 4,900 images captured under low-light conditions, taken from five unique angles. It comprises 3,920 images in the training set and 980 in the testing set. The KDEF dataset \cite{kdef}, with an identical total of 4,900 images, is a comprehensive collection in which each facial expression is photographed from five distinct viewpoints, ensuring a broad spectrum of clear visual information.
\subsection{Implementation Details}
We use Adobe Lightroom \cite{kosugi2020unpaired} to create three benchmark low-light facial expression recognition (LL-FER) datasets, simulating degraded image conditions by adjusting the exposure, white balance, highlights, and shadows. The experimental setup utilized PyTorch for model training, which was carried out on a GTX-3090 GPU. For optimization, the Adam algorithm was chosen, with the training spanning 200 epochs. The adopted training settings specified an initial learning rate of $3.5 \times 10^{-4}$, a batch size of 64,  and a weight decay parameter set to $1 \times 10^{-4}$.
\subsection{Comparison with other SOTA FER Methods}
\textbf{Comparison with other Typical state-of-the-art FER Methods.} 
Table \ref{three-datasets-sota-comparison} offers a detailed evaluation of the accuracy of our proposed approach against the latest SOTA facial emotion recognition techniques \cite{2020scn,2021manet,2021arm,2022eac,2021dacl,2021rul,2023dan,2023posterv2} over three standard FER datasets: RAF-DB, KDEF and FERPlus. For RAF-DB, our method records a 91.72\% accuracy, outperforming several well-established algorithms such as  RUL\cite{2021rul}, ARM\cite{2021arm}, DAN\cite{2023dan}, EAC\cite{2022eac}, SCN\cite{2020scn} and MANet\cite{2021manet}, and is closely matched with POSTER++\cite{2023posterv2} which has a marginally higher accuracy of 92.21\%. On the FERPlus dataset, the proposed method demonstrates an 87.19\% accuracy, exceeding the accuracy of RUL \cite{2021rul} at 85.00\%, POSTER++ \cite{2023posterv2} at 86.46\%, EAC\cite{2022eac} at 86.18\%, and MANet \cite{2021manet} at 85.49\%, and the SCN\cite{2020scn} method has a lowest performance compared to the other methods. Within the KDEF dataset analysis, our proposed approach secures the top accuracy at 95.83\%, showcasing a progress against other approaches, surpassing POSTER++\cite{2023posterv2} at 94.44\% and MANet\cite{2021manet} at 91.75\%. Overall, these results underscore the reliability of the proposed approach in handling facial expression recognition across various datasets.

\begin{figure}[ht!]
\centering
\includegraphics[width=0.98\linewidth]
{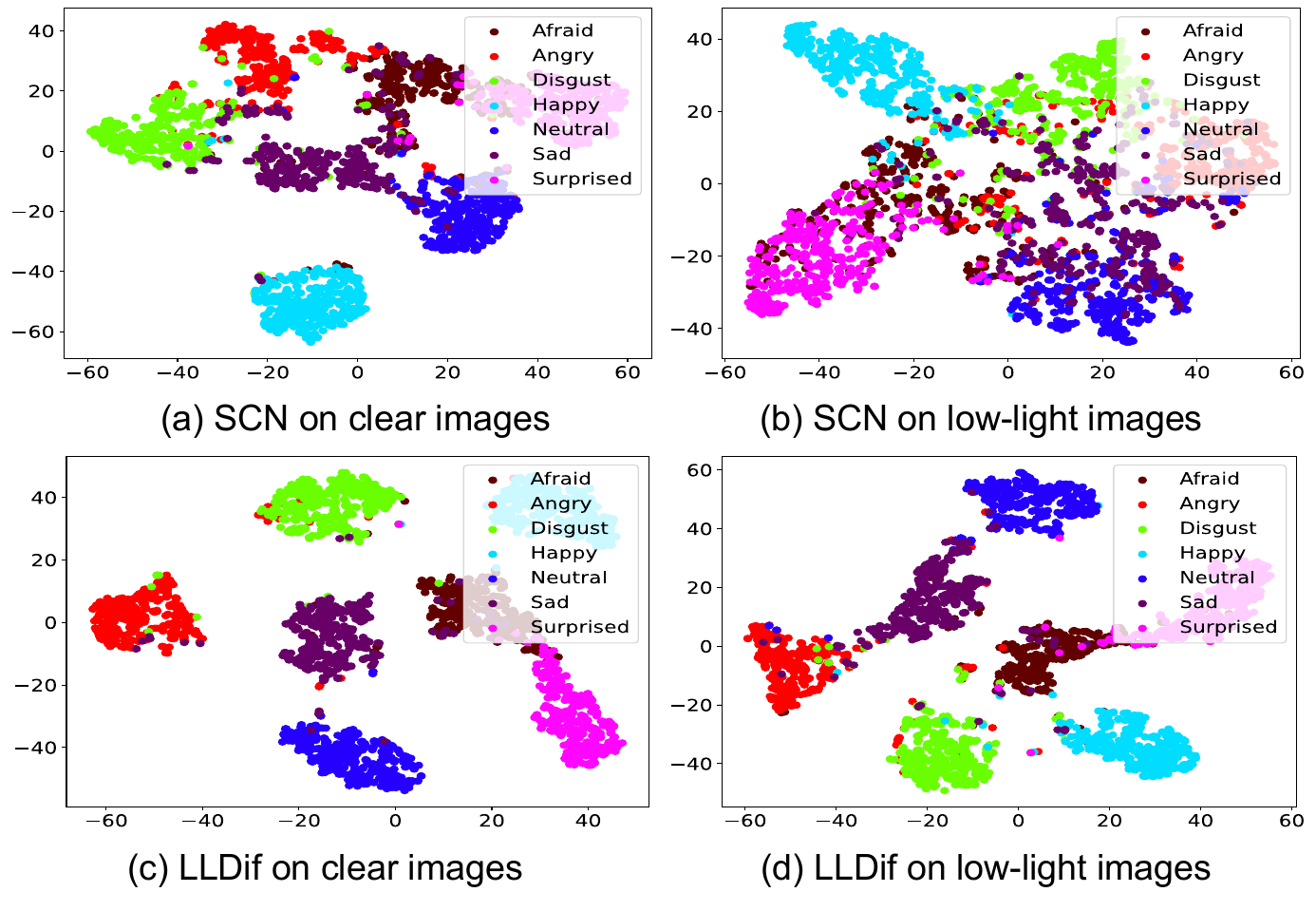}
\caption{The predicted feature visualised by t-SNE between our method and SCN.}
\label{tsne_scn_ours_comparision}
\end{figure}
\begin{table}[ht!]
\centering
\smallerfonttable 
\setlength{\tabcolsep}{8.8pt} 
\renewcommand{\arraystretch}{1} 
\caption{Key components in LLDif.}
\label{ablation_study_components}
\renewcommand{\arraystretch}{1.2} 
\begin{tabular}{l|ccccc}  
\toprule[1pt] 
\multirow{2}{*}{\textbf{Method}} & \multicolumn{4}{c}{\textbf{Key components in LLDif}} &\multirow{2}{*}{\textbf{Acc.(\%)}} \\ \cmidrule(l){2-5} 
 & \textbf{Diffusion Model} &\textbf{$\mathcal{L}_{ce}$} & \textbf{$\mathcal{L}_{total}$}& \textbf{Insert Noise} & \\
\midrule 
LLDif$_{S2}$-V1    &  \XSolidBrush  &\CheckmarkBold& \XSolidBrush  & \XSolidBrush & 89.46\\
LLDif$_{S2}$-V2 & \CheckmarkBold  &\XSolidBrush & \CheckmarkBold  & \CheckmarkBold  & 91.67\\
LLDif$_{S2}$-V3 &\CheckmarkBold &\CheckmarkBold  & \XSolidBrush &  \XSolidBrush  & 92.16 \\
LLDif$_{S2}$-V4 (Ours) & \CheckmarkBold & \CheckmarkBold  & \XSolidBrush  &  \CheckmarkBold & 92.97\\
\bottomrule[1pt] 
\end{tabular}
\end{table}
\textbf{Comparison with the low-light FER-model.} We compare our method with other SOTA methods on low-light images. Some samples are shown in Fig. \ref{probability-distribution-ours}. Accuracy comparisons between our model and other SOTA FER methods on the LL-RAF-DB, LL-KDEF datasets, and LL-FERPlus  are outlined in Tables \ref{udc-raf-db-sota-comparison}, \ref{udc-ferplus-sota-comparison}, and \ref{udc-kdef-sota-comparison}, respectively. The majority of the benchmarked models, including ARM\cite{2021arm}, RUL\cite{2021rul}, DAN\cite{2023dan}, SCN\cite{2020scn}, EAC\cite{2022eac}, and MANet\cite{2021manet}, are based on the ResNet-18 architecture. However, POSTER++\cite{2023posterv2} stands out by adopting the Vision Transformer architecture. In contrast, our model introduces a novel approach by incorporating a 'Diffusion' backbone, moving away from the traditional ResNet-18 design. In the Table \ref{udc-raf-db-sota-comparison}, the proposed method attains the highest accuracy of 82.26\%, which is a notable enhancement over other methodologies. POSTER++ \cite{2023posterv2} registers the second highest accuracy with 80.76\%, followed by DAN \cite{2023dan}at 79.27\%. The EAC\cite{2022eac}, MANet\cite{2021manet}, RUL\cite{2021rul}, SCN\cite{2020scn}, and DACL\cite{2021dacl} algorithms show a relative low accuracy from 75.20\% to 78.72\%. Table \ref{udc-ferplus-sota-comparison}, which focuses on the LL-FERPlus Dataset, shows "Ours" with a leading accuracy of 82.25\%, marginally surpassing POSTER++'s \cite{2023posterv2} 81.44\%. In the Table \ref{udc-kdef-sota-comparison}, "Ours" shows the highest accuracy at 92.97\%, which is significantly higher than the other methods listed. The second most accurate method is POSTER++\cite{2023posterv2}, with an accuracy of 88.93\%. Other methods such as DAN\cite{2023dan}, EAC\cite{2022eac}, MANet\cite{2021manet}, RUL\cite{2021rul}, SCN\cite{2020scn}, and DACL\cite{2021dacl} present accuracies ranging from 43.53\% to 86.69\%. These results underscore the efficiency of the diffusion-based approach within the context of facial expression recognition systems under low-light conditions.

\begin{figure*}[t!]
\centering
\includegraphics[width=0.95\linewidth]{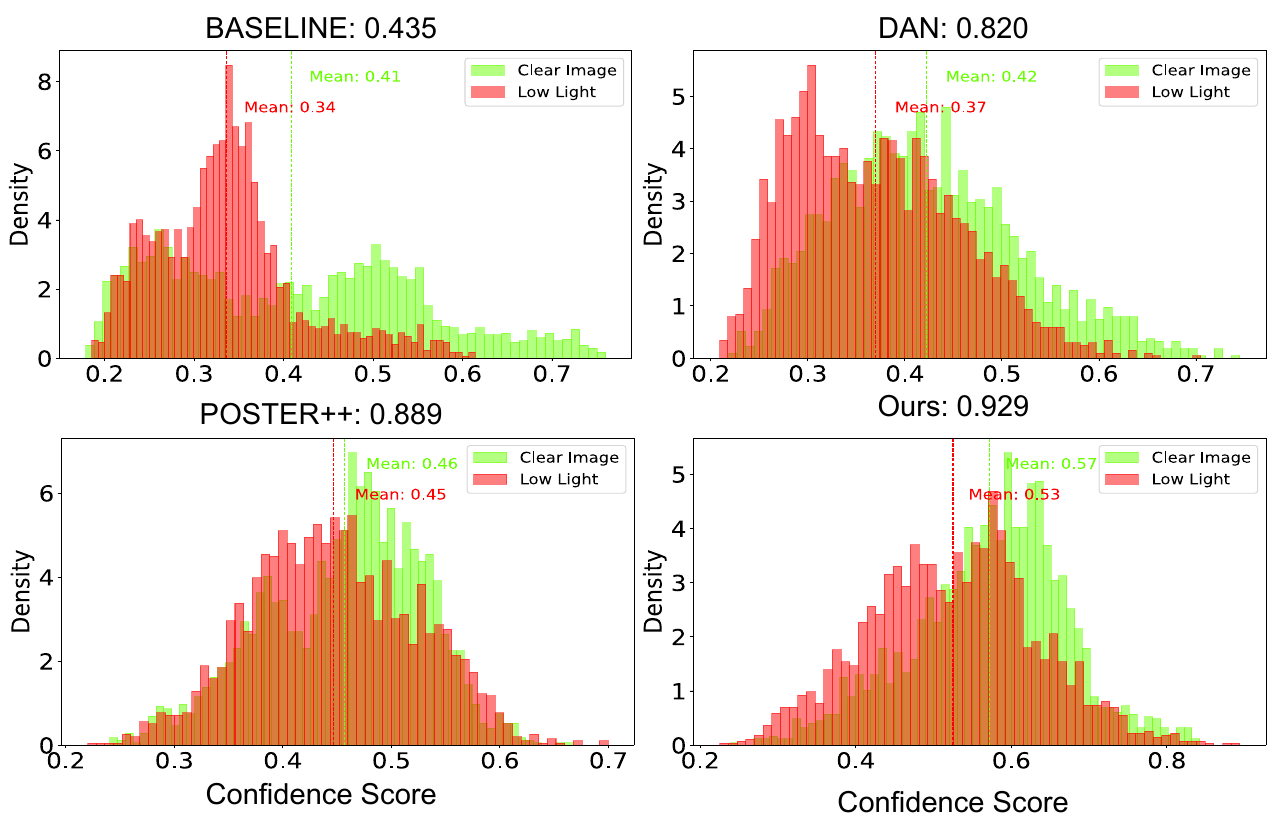}
\caption{Confidence score of different methods on KDEF dataset. Accuracy for each method is marked on the top. The baseline \cite{2020scn} method fails as FER data have small inter-class distances. DAN\cite{2023dan} and POSTER++\cite{2023posterv2} have relative high confidence score while they still fall a lot in low confidence score area. Our method can effectively separates different emotion samples on clear and low-light images.}
\label{confidence_score}
\end{figure*}
\textbf{Feature Visualization.} We used the t-SNE method to illustrate how models discern feature distributions. In contrast to Fig. \ref{tsne_scn_ours_comparision} (a) and (b), where the SCN model has difficulty separating different emotion categories, especially in low-light conditions, our LLDif model exhibits effective expression recognition in both clear and degraded low-light images. This indicates that LLDif successfully captures key features crucial for distinguishing between various emotional expressions categories.

\textbf{Visualization of Confidence Scores.} We visualize the distribution of confidence scores for facial expression recognition methods on clear and low-light images in Fig. \ref{confidence_score}. For the baseline method \cite{2020scn}, the mean confidence score for clear images is 0.41 and for low-light images is 0.34, with an overall accuracy of 0.435. The DAN method \cite{2023dan} shows a mean confidence score of 0.42 for clear images and 0.37 for low-light images, with an overall accuracy of 0.820. The POSTER++ method\cite{2023posterv2} has mean scores of 0.46 for clear images and 0.45 for low-light images, achieving an overall accuracy of 0.889. The proposed method exhibits a notably higher confidence level with mean scores of 0.57 for clear images and 0.53 for low-light images, corresponding to a high overall accuracy of 0.929. The proposed method not only shows the highest accuracy but also the small difference in confidence score between clear and low-light images, suggesting robust performance even in challenging lighting conditions.
\begin{figure}[t!] 
\centering
\begin{subfigure}{0.24\linewidth}
\caption{T = 1 }
\includegraphics[width=\linewidth]{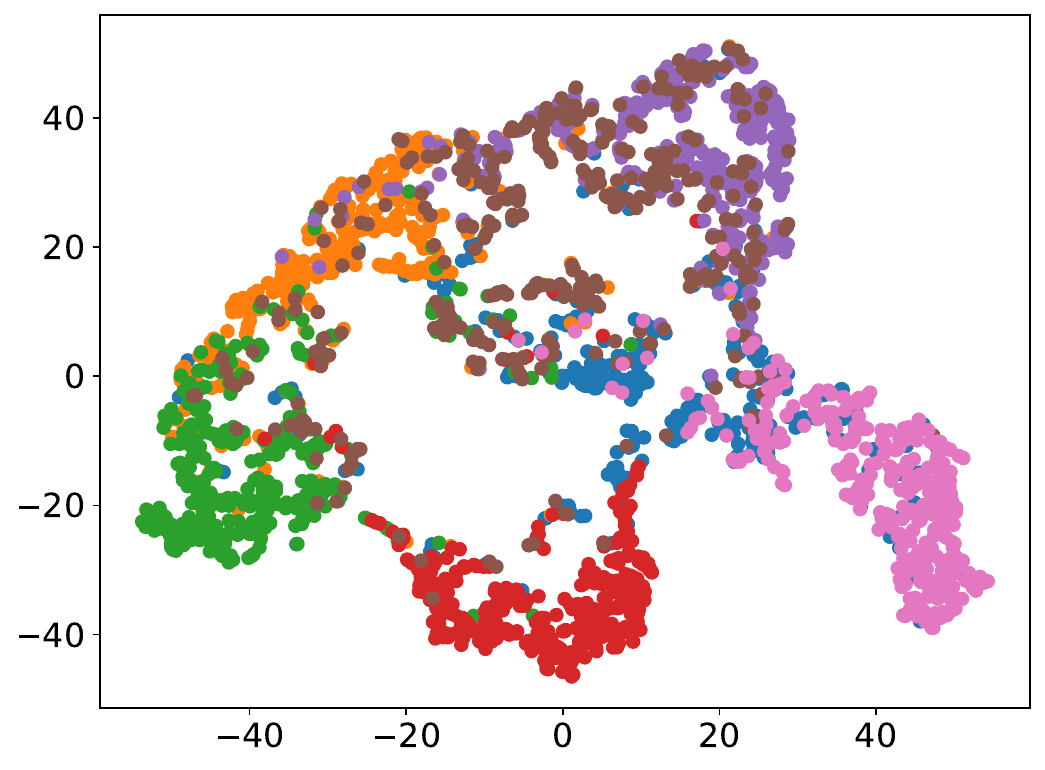}
\end{subfigure}
\begin{subfigure}{0.24\linewidth}
\caption{T = 2 }
\includegraphics[width=\linewidth]{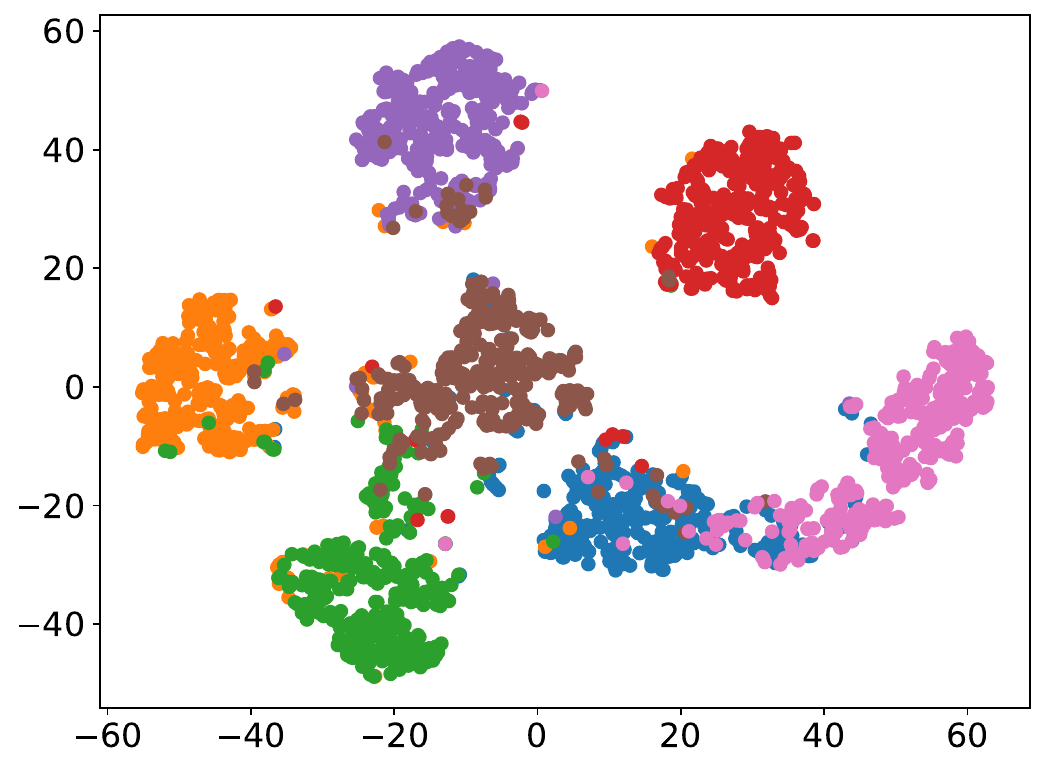}
\end{subfigure}
\begin{subfigure}{0.24\linewidth}
\caption{T = 3}
\includegraphics[width=\linewidth]{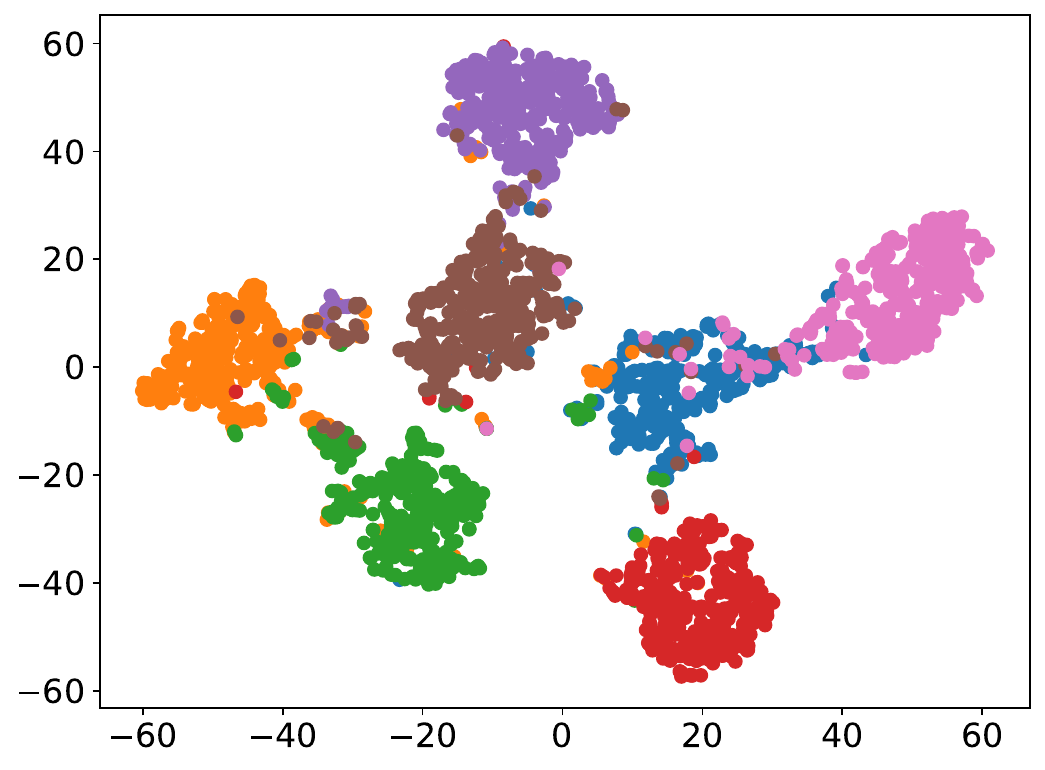}
\end{subfigure}
\begin{subfigure}{0.24\linewidth}
\caption{T= 4}
\includegraphics[width=\linewidth]{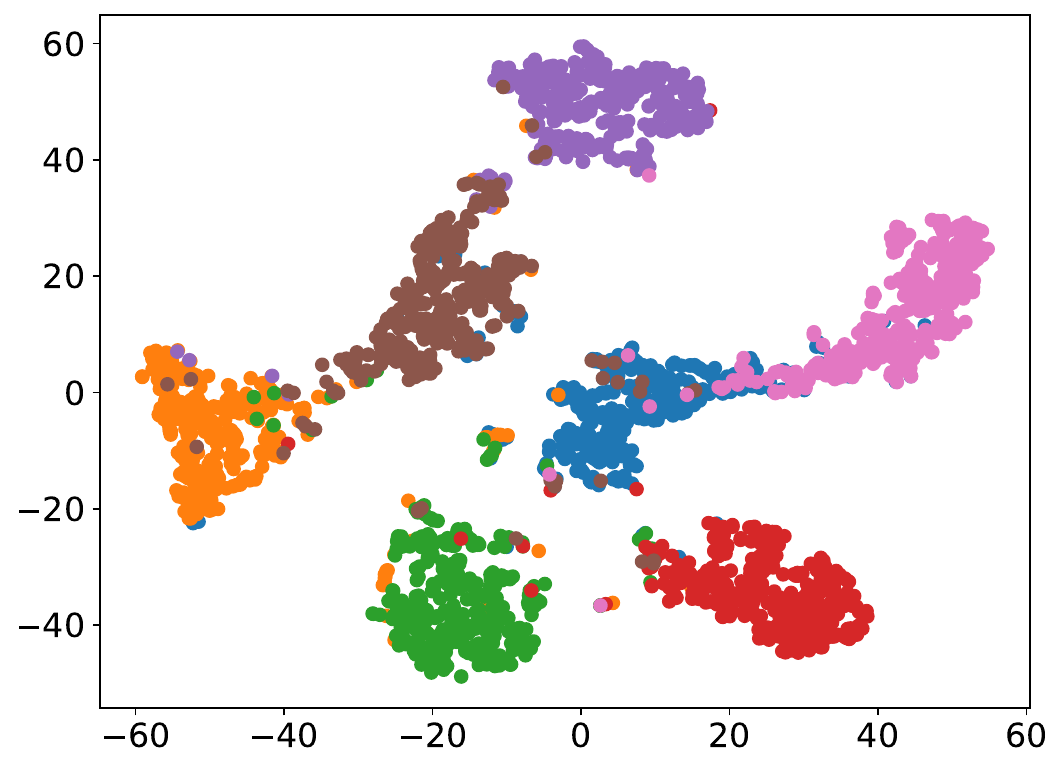}
\end{subfigure}
\begin{subfigure}{0.24\linewidth}
\caption{T= 6}
\includegraphics[width=\linewidth]{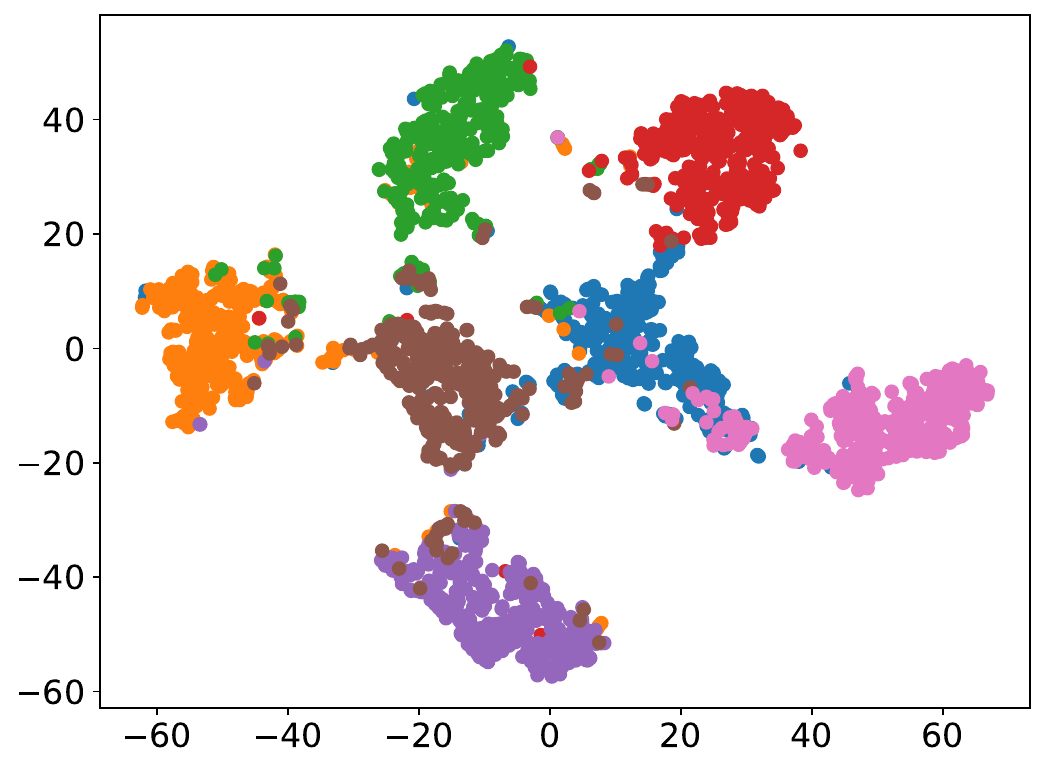}
\end{subfigure}
\begin{subfigure}{0.24\linewidth}
\caption{T= 10}
\includegraphics[width=\linewidth]{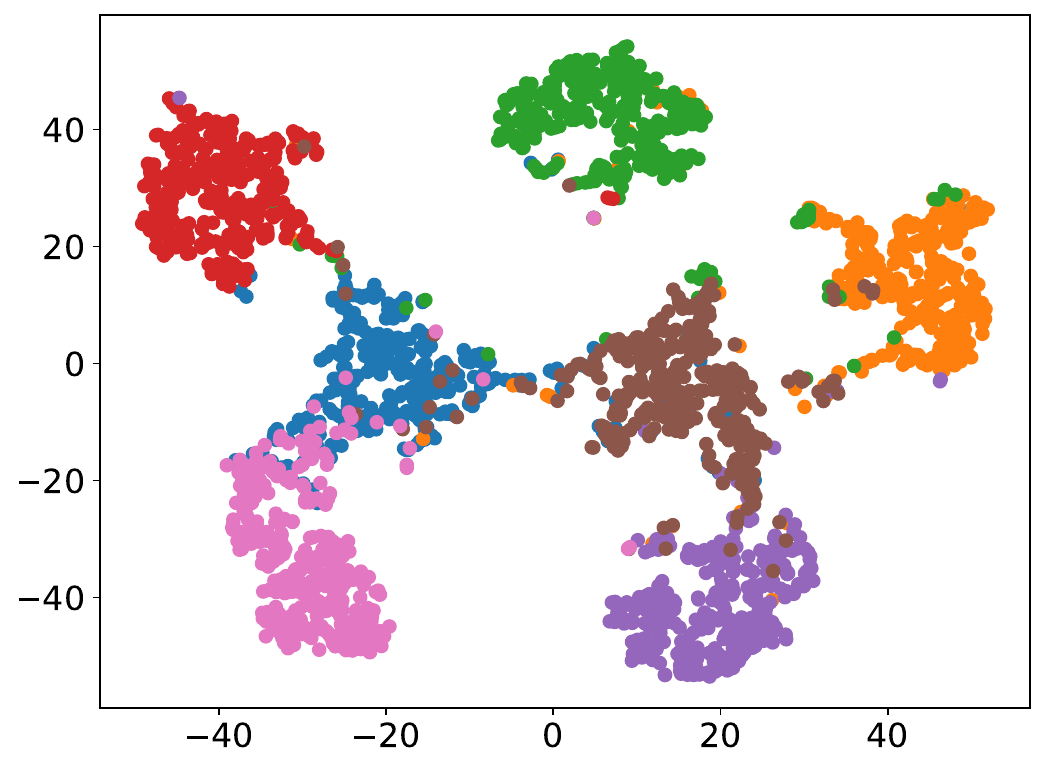}
\end{subfigure}
\begin{subfigure}{0.24\linewidth}
\caption{T= 20}
\includegraphics[width=\linewidth]{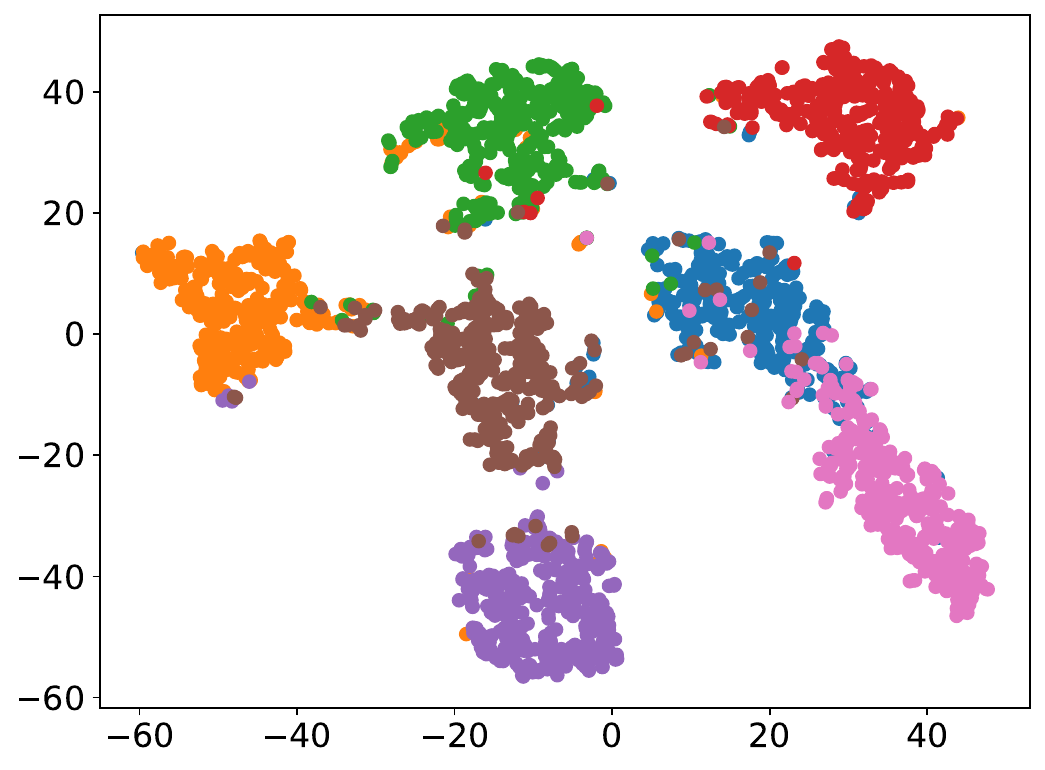}
\end{subfigure}
\begin{subfigure}{0.24\linewidth}
\caption{T= 32}
\includegraphics[width=\linewidth]{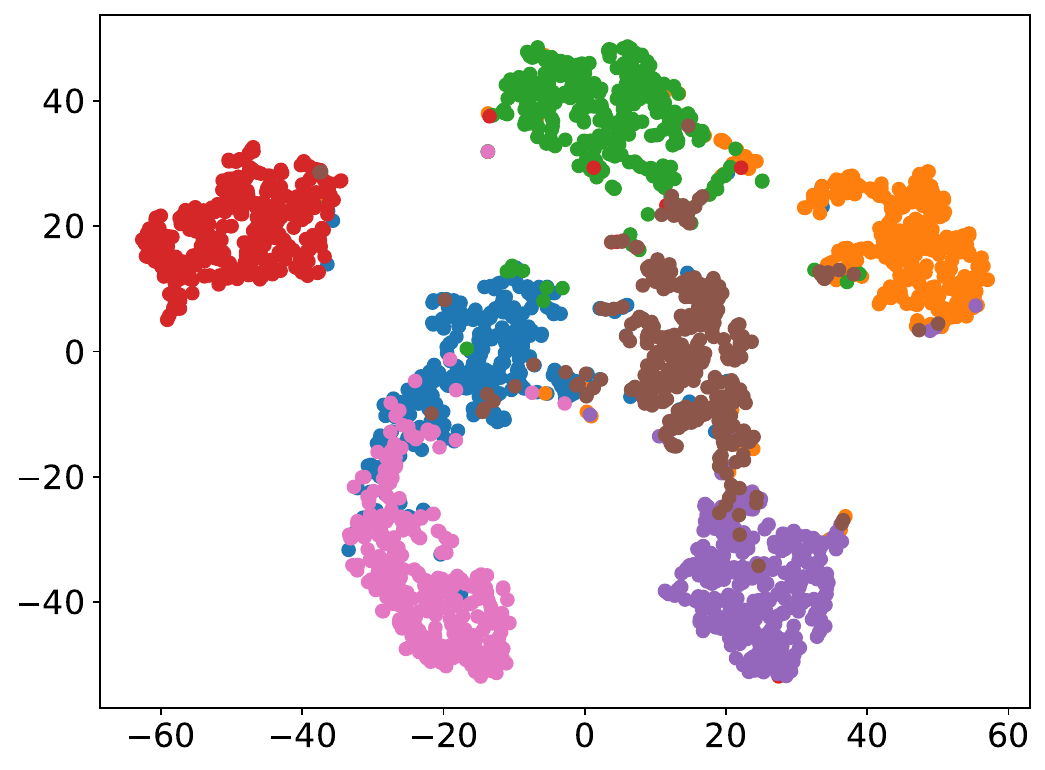}
\end{subfigure}
\caption{Progressive clustering of features in diffusion space visualized using t-SNE at different time steps (T).}
\label{t-sne-feature-DDPM}
\end{figure}
\begin{figure}[t!]
\centering
\includegraphics[width=0.8\linewidth]{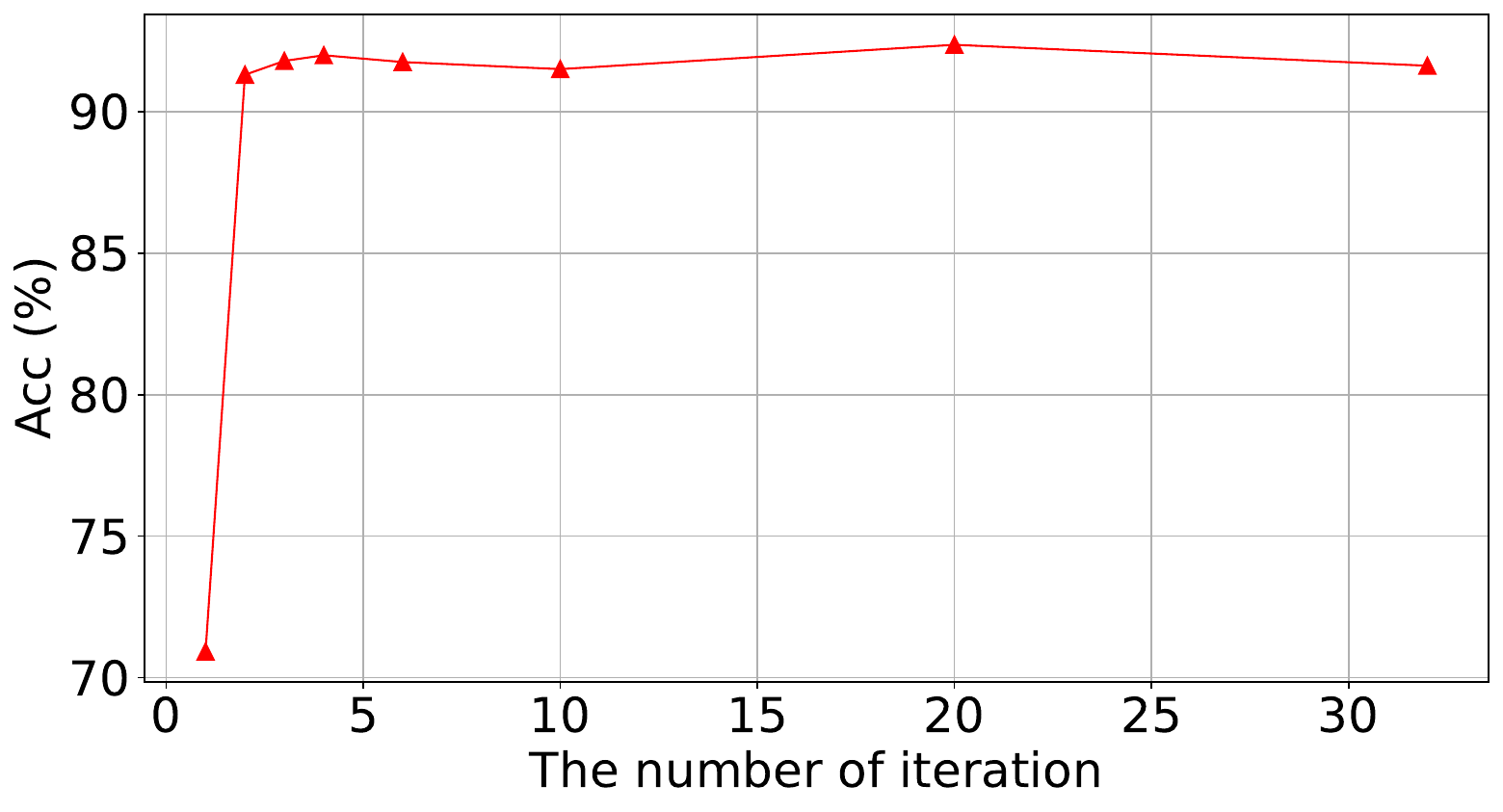}
\caption{Analyse impacts of iterations in DM.}
\label{number_of_iterations}
\end{figure}

\subsection{Ablation Study}
This section evaluates the impacts of crucial components within LLDif, including the Diffusion Model (DM), various loss functions, and the insert noise during the training phase, as depicted in Table \ref{ablation_study_components}. (1) The contrast between LLDif$_{S2}$-V3 and LLDif$_{S2}$-V1 underscores the DM's robust ability in accurately predicting the embedding prior distribution EPD. (2) The insert noise into the DM's process in LLDif$_{S2}$-V4 is demonstrated to enhance the accuracy of EPD predictions. (3) The efficiency of different loss functions is also examined. The comparison between using $\mathcal{L}_{ce}$ in LLDif$_{S2}$-V4 (refer to Eq. (\ref{l_ce_loss})) and $\mathcal{L}_{total}$ in LLDif$_{S2}$-V2 (refer to Eq. (\ref{l_total_loss})) shows that using $\mathcal{L}_{ce}$ is required for achieving better accuracy.

\textbf{Impact of iteration numbers.} 
This section examines how varying the number of iterations in the Diffusion Model (DM) influences the LLDif$_{S2}$ performance. We experimented with different iteration numbers in LLDif$_{S2}$, adjusting the $\beta_t$ value (with $\alpha_t$ set as $1 - \beta_t$, as outlined in Eq. \ref{sample_z}) to ensure the variable $Z$ evolves toward a Gaussian distribution, $Z_T \sim \mathcal{N}(0,1)$. Figures \ref{number_of_iterations} and \ref{t-sne-feature-DDPM} demonstrate that LLDif$_{S2}$'s performance notably enhances at 4 iterations. Increasing the iteration number over 4 iterations does not substantially impact model's performance, suggesting the attainment of an optimal threshold. Notably, LLDif$_{S2}$ reaches convergence more quickly than traditional DM methods, which typically requires over 50 iterations. This enhanced efficiency results from applying DM on the EPD, which is a one-dimensional, concise vector.
\section{Conclusion}
In this work, we present LLDif, an innovative framework utilising diffusion-based method to enhance facial expression recognition under low-light conditions. Addressing the challenges of image quality degradation in low-light settings, LLDif employs a two-stage training approach, utilizing a label-aware CLIP (LA-CLIP), an embedding prior distribution network (PNET), and a diffusion-based transformer network (LLformer). By integrating advanced architecture like the PNET and LLformer,  LLDif can effectively restore emotion labels from degraded low-light images at multiple scale. Our experiments confirms that LLDif outperforms existing methods, gains competitive performance on three low-light facial expression recognition datasets.
%
%
%
\bibliographystyle{splncs04}
\bibliography{refs_low_light}
%




\end{document}